\documentclass[acmsmall]{acmart}

\AtBeginDocument{%
  }

\setcopyright{acmlicensed}
\copyrightyear{2018}
\acmYear{2018}
\acmDOI{XXXXXXX.XXXXXXX}

\acmJournal{TOMM}
\acmVolume{37}
\acmNumber{4}
\acmArticle{111}
\acmMonth{8}
\usepackage[capitalize]{cleveref}
\usepackage[ruled,linesnumbered]{algorithm2e}
\usepackage{multirow}

\begin{document}

\title{From Recognition to Prediction: Leveraging Sequence Reasoning for Action Anticipation}


\author{Xin Liu}
\authornote{Both authors contributed equally to this research.}
\email{linuxsino@gmail.com}
\affiliation{%
  \institution{School of Electrical and Information Engineering, Tianjin University
  }
  \city{Tianjin}
  \country{China}
  \institution{, and Computer Vision and Pattern Recognition Laboratory, School of Engineering Sciences, Lappeenranta-Lahti University of Technology LUT}\city{Lappeenranta}  \country{Finland}
}

\author{Chao Hao}
\authornotemark[1]
\email{3018234336@tju.edu.cn}
\affiliation{%
  \institution{School of Electrical and Information Engineering, Tianjin University}
  \city{Tianjin}
  \country{China}
}

\author{Zitong Yu}
\authornote{Corresponding author.}
\affiliation{%
  \institution{School of Computing and Information Technology, Great Bay University}
  \city{Dongguan}
  \country{China}}
\email{zitong.yu@ieee.org}

\author{Huanjing Yue}
\email{huanjing.yue@tju.edu.cn}
\author{Jingyu Yang}
\email{yjy@tju.edu.cn}
\affiliation{%
  \institution{School of Electrical and Information Engineering, Tianjin University}
  \city{Tianjin}
  \country{China}
}

\renewcommand{\shortauthors}{Liu et al.}

\begin{abstract}
 The action anticipation task refers to predicting what action will happen based on observed videos, which requires the model to have a strong ability to summarize the present and then reason about the future. 
Experience and common sense suggest that there is a significant correlation between different actions, which provides valuable prior knowledge for the action anticipation task. However, previous methods have not effectively modeled this underlying statistical relationship. To address this issue, we propose a novel end-to-end video modeling architecture that utilizes attention mechanisms, named Anticipation via Recognition and Reasoning (ARR). ARR decomposes the action anticipation task into action recognition and sequence reasoning tasks, and effectively learns the statistical relationship between actions by next action prediction (NAP).
 In comparison to existing temporal aggregation strategies, ARR is able to extract more effective features from observable videos to make more reasonable predictions. 
 In addition, to address the challenge of relationship modeling that requires extensive training data, we propose an innovative approach for the unsupervised pre-training of the decoder, which leverages the inherent temporal dynamics of video to enhance the reasoning capabilities of the network.
 Extensive experiments on the Epic-kitchen-100, EGTEA Gaze+, and 50salads datasets demonstrate the efficacy of the proposed methods. The code is available at  \href{https://github.com/linuxsino/ARR}{https://github.com/linuxsino/ARR}. 
\end{abstract}



\received{20 February 2007}
\received[revised]{12 March 2009}
\received[accepted]{5 June 2009}

\maketitle

\section{Introduction}
Imagine a future world where a smart home robot is taking care of a child, suddenly, the robot senses that the child is showing signs of falling and helps to support the child in time to avoid potential harm to the child. This proactive intervention to prevent falls is more important than simply helping a child get up after a fall. Predicting the future actions of humans is an important and interesting task for artificial intelligence systems. It is currently receiving widespread attention \cite{AVT, AFFT, RULSTM, actionfore} and has many applications in our lives. If self-driving cars can avoid pedestrians in advance, surveillance videos can detect certain dangers and forewarning, these technologies will make our lives better.
    
\begin{figure}[t]
      \centering
      \includegraphics[width=0.8\linewidth]{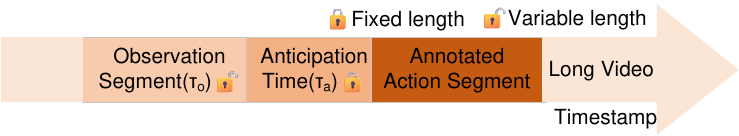}
      \vspace{-3mm}
      \caption{Description of action anticipation task. In a long video, observed segments of length  $ \tau_{o}$  are used to infer actions after an anticipation time of length $ \tau_{a}$. “Observation Segment” is used as input, and the output is the action category of “Annotated Action Segment” that we predict, and then calculates the prediction loss with its corresponding annotation.  $ \tau_{o}$ is not fixed and can be selected according to needs, while $ \tau_{a}$ is fixed. Different datasets have different settings.}
      \label{fig:description}
      \vspace{-4mm}
\end{figure}

The current general action anticipation task refers to predicting the (verb, noun) label of a future action by observing the segment before it occurs \cite{EK55,EK100+,EK100}, as shown in \cref{fig:description}. Action anticipation is similar to the action recognition task \cite{tomm1, tomm2}, both tasks take a video as input and produce an action classification as output. However, the former predicts future actions, while the latter recognizes actions in the present. These two tasks typically utilize the same datasets and annotations \cite{EK100, gteadataset, 50salads}. In a lengthy video, the action recognition task solely employs labeled action segments. Conversely, the input of action anticipation comprises both labeled action segments and unlabeled segments, with variable input length. Evidently, action anticipation task is considerably more challenging, with certain state-of-the-art action recognition models demonstrating subpar performance in action anticipation task \cite{tar, tsm}.

\begin{figure}[t]
      \centering
      \vspace{-3mm}
      \includegraphics[width=0.8\linewidth]{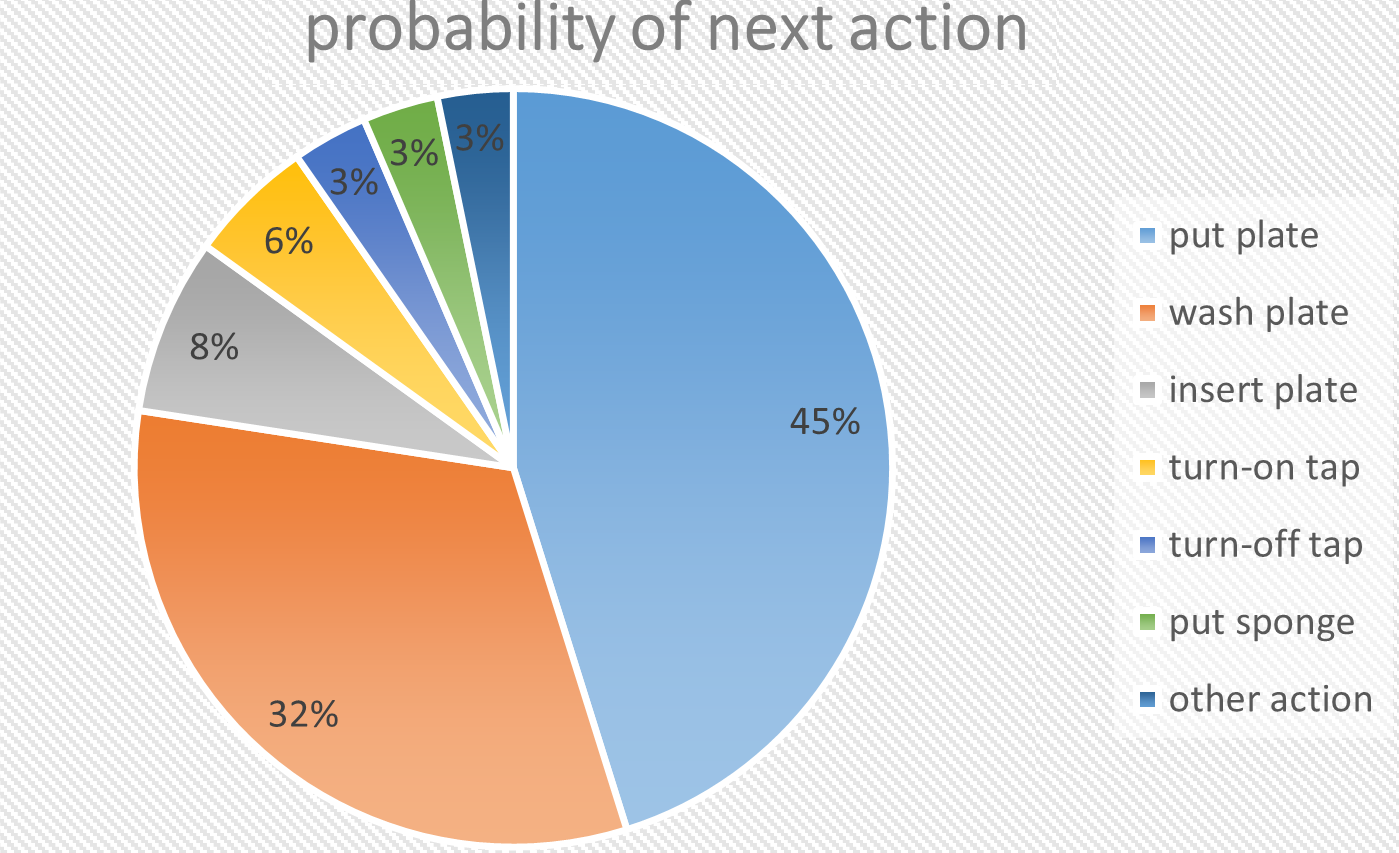}
      \caption{The probability of next action. We randomly select an action in EK100 dataset as the current action (e.g., ``wash plate''). We count the frequency of the next action, and use the frequency to calculate the probability. Although there are thousands of actions in the dataset, we can see that only some actions (e.g., ``put plate'' and ``wash plate'') highly correlated will occur after the given action, and most actions will not appear.}
      \label{fig:probability}
      \vspace{-4mm}
   \end{figure}

When making predictions, humans typically assess the current situation first and then draw on past experiences to inform their judgments. For example, if someone is chopping vegetables in the kitchen, it's more likely that he will cook rather than cleaning. Clearly, based on our experience, different actions are interconnected, and there is a statistical relationship between the likelihood of their occurrence. This relationship is illustrated in \cref{fig:probability}. We analyze the frequency of pairs of actions, consisting of a randomly chosen action followed by a subsequent action. Our findings show that only a few types of actions typically follow any given action. According to our experience, these sequences have strong correlations. The majority of potential subsequent actions are unlikely to occur after the initial action, providing crucial prior information that is beneficial for the task of action anticipation. However, previous methods often ignored this important factor.
   
\begin{figure}[t]
      \centering
      \vspace{-4mm}
      \includegraphics[width=0.7\linewidth]{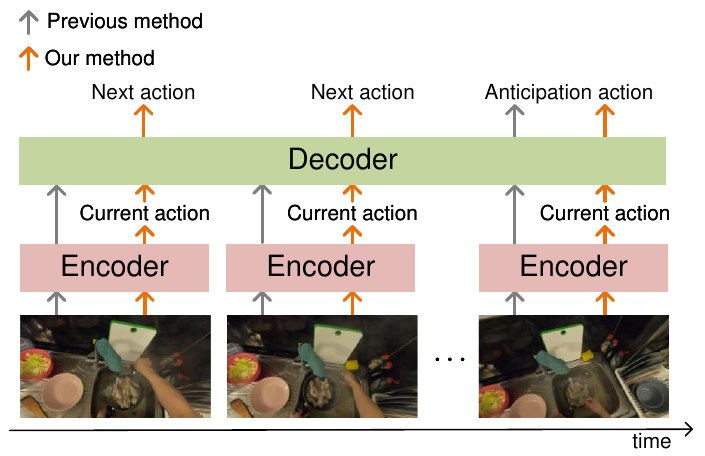}
      \caption{Comparison of our method with previous methods. While previous methods tend to predict actions directly from a given video, our method learns the correlation among actions to benefit prediction by recognizing actions and then conducting sequence reasoning.}
      \vspace{-4mm}
      \label{fig:comparison}
   \end{figure}

The conventional strategy for addressing long-term predictive reasoning tasks entails initial feature extraction at the frame or clip level \cite{k400dataset, 3dcnn} using established architectures. Subsequently, aggregation is achieved through various methods, including clustering \cite{aggregation}, recurrence \cite{recurrence2, recurrence1}, or models based on attention mechanisms \cite{attention1, attention2}. They are more about iterating and aggregating the extracted features, while ignoring the deeper correlation among actions for action anticipation. In contrast, we think that learning the intrinsic correlations among different actions is beneficial in making more reasonable predictions. As shown in \cref{fig:comparison}, our method will first get the current action and then go through the action sequence to get the final result instead of directly getting the prediction like the previous method.

In this work, we propose Anticipation via Recognition and Reasoning (ARR). We decompose the action anticipation task into two tasks of action recognition and sequence reasoning, which have been studied a lot. We model the input video as a sequence input, and map them into action sequences through action recognition task. Then we use the causal decoder structure that has been proven to have powerful reasoning capabilities by large language models \cite{gpt1, gpt2, gpt3, lamda, llama, opt, bloom} to complete sequence reasoning task. We draw on the popular Next Token Prediction (NTP) task \cite{gpt1} of the current large language model and propose Next Action Prediction (NAP), which enables the network to fully learn and model the correlation among different actions, and enhance the reasoning ability of the network. In addition, to solve the problem that NAP requires a large amount of training data, we also use an unsupervised pre-training method \cite{unsupervised, AVT, tomm4} to initialize the sequence reasoning network to further enhance the performance of the network.

Specifically, ARR uses vision Transformer \cite{vit, timesformer, vivit, senet} and causal decoder \cite{transformer} to realize the main function. Since the emergence of vision Transformer, action recognition technology has developed very advanced. The highest recognition accuracy of the K400 dataset \cite{k400dataset} has reached $91\%$ \cite{internvideo}, which has exceeded the human level. Using action recognition task for supervised training can ensure that we extract more effective features from observation clips \cite{videoswin}, which pay more attention to action content rather than other useless information. In addition, causal masked self-attention ensures that each predicted action can only obtain information from its front, and there is no risk of information leakage, which is crucial for the action anticipation task \cite{predicctfuture, EK100}. Through this paradigm, the network can be based on actions that occurred in the past to predict the next action and realize the modeling of the correlation between actions. As for the unsupervised pre-training based on the natural timing of the video before the overall training, we do not change the overall architecture of the network, but replacing the supervised learning method with the self-supervised form of learning the next feature.

In summary, our main contributions include: 

\begin{itemize}
    \item We propose the ARR, a novel end-to-end architecture for implementing action anticipation, which is based on the pure attention mechanism and effectively completes the action anticipation task.

    \item We decompose the relatively difficult task of action anticipation into two simpler tasks (i.e., action recognition and sequence reasoning), and propose the NAP to learn the intrinsic correlation among different actions. Thus, the model can better utilize the statistic relationship among actions to complete the inference.

    \item We propose a method for unsupervised pre-training using the natural timing of videos, which provides an effective method for network initialization of action anticipation task. 

    \item We achieve competitive results on the Epic-kitchen-100, EGTEA Gaze+ and 50salads datasets and demonstrate a strong statistical correlation among actions.
\end{itemize}

\section{Related Work}
\noindent \textbf{Action Recognition} is essentially a classification task of judging actions from a given video. 
In the early days, people extracted features through manual design methods \cite{traditional1, traditional2, traditional3, tomm3}, these methods often require careful design but the performance is not good enough. With the emergence of deep learning methods \cite{AlexNet}, the accuracy of action recognition has begun to increase significantly, and various efficient algorithms and large-scale datasets have also begun to appear. Karpathy et al. \cite{cnn_rec} considered the use of convolutional neural networks (CNN) and investigated different strategies for fusing per-frame predictions. Simonyan et al. \cite{twostream} proposed two-stream CNN, a multi-branch architecture for action recognition by processing appearance (RGB) and motion (optical flow) data.  Carreira et al. \cite{k400dataset} proposed to expand 2D CNN into 3D to make better use of large-scale pre-training of images. 
Timesformer \cite{timesformer} was the first to propose a video-based Transformer architecture and achieved good results, and then the attention-based methods directly dominated this field \cite{ videoswin, uniformerv2, unmaskedvideo, memvit}. The action recognition module in this work is also based on Transformer architecture \cite{aim}.

\begin{figure}[t]
    \centering
      \includegraphics[width=\linewidth]{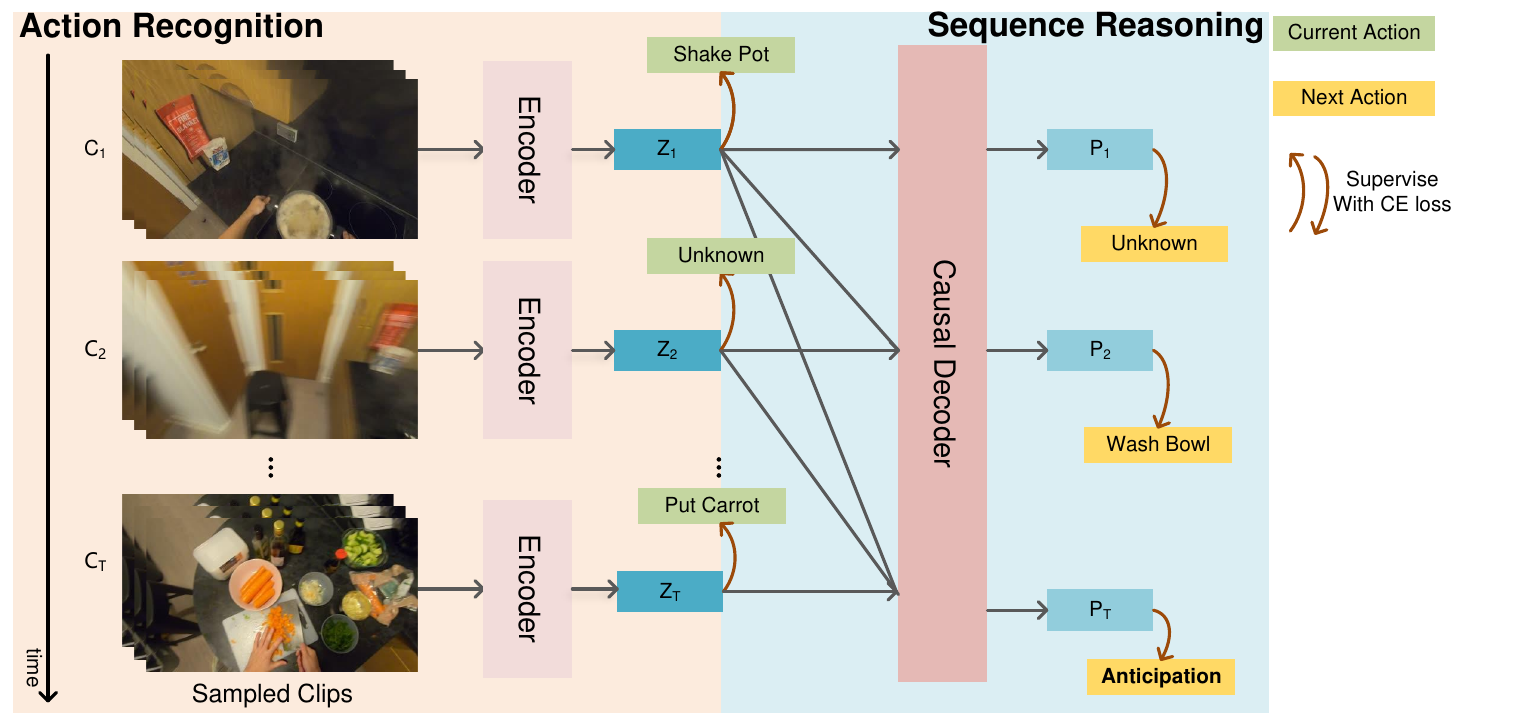}
      \vspace{-3mm}
      \caption{Overall framework of the ARR. The model consists of two components: on the left is a complete action recognition network that identifies the action corresponding to the current video clip within the action annotation sequence. On the right is a complete sequence reasoning network that predicts the next action corresponding to the current video clip within the action annotation sequence.}
      \vspace{-4mm}
      \label{fig:Architecture}
\end{figure}

\vspace{1mm}
\noindent \textbf{Action Anticipation} task \cite{tomm5} appeared not long ago but has received widespread attention, especially in the field of first-person video. The current mainstream method is to sample the video to form a sequence, then extract information through the encoder, and finally predict the anticipation action by aggregating the extracted features through LSTM \cite{lstm} or other sequence models\cite{gru, gpt2}. Furnari et al. \cite{RULSTM} used pre-extracted frame-level features for recursive modeling and prediction via LSTM \cite{lstm}. Girdhar et al. \cite{AVT} first used the pre-trained causal decoder as the prediction head to greatly improve the prediction ability. Zhong et al. \cite{AFFT} explored the relationship between different modal information and the influence of the interaction mode between them on the prediction results. This work is also based on the encoder and decoder structure but has been modified for some specific problems. Roy et al. \cite{inavit} proposed a new strategy that greatly improved the indicators of action anticipation tasks by paying special attention to human-object interaction and temporal dynamics along the motion paths.

Action recognition and action prediction are two highly related tasks, one focuses on the present and the other focuses on the future. There have been some works \cite{peeking, liang2020recognition} trying to jointly learn them. The method proposed in this paper is to predict the future on the basis of a better understanding of the present, which is a reasonable and effective extension of end-to-end action recognition to action prediction.

\vspace{1mm}
\noindent \textbf{Next Token Prediction (NTP)} is the essence of text generation and a key to the success of large language models \cite{gpt3}. It has almost become the basic pre-training task of all large language models \cite{palm, opt, bloom, llama, gpt3, gopher}. The language model learns the potential relationship between languages by predicting the probability of the next word. The GPT series were the first to use this NTP-based task to train a model, it fits perfectly with the pure causal decoder structure. Although the performance was not as good as Bert \cite{bert} at the beginning, with the increase of training data, GPT3.5 made people see the hope of powerful artificial intelligence. In this work, we follow the example of NTP to construct the Next Action Prediction (NAP), hoping that the model can better learn the statistical relationship among actions through this training method. 
NAP predicts the next action based on a sequence of preceding actions. Initially, a video encoder is required to convert input video clips into an action sequence, followed by the use of a decoder to model the relationships between actions.

\vspace{1mm}
\noindent \textbf{Self-supervised pre-training.} The model capability of supervised learning is always limited by the size of the dataset \cite{mae}. In some cases the labeled dataset is too expensive and requires a lot of manpower and material resources, so people have been hoping to construct some self-supervised learning tasks to overcome the limitation of a lack of training data. Masked Autoencoder (MAE) \cite{mae} imitates Bert's \cite{bert} cloze task to obtain the ability to extract effective features by masking and reconstructing pictures. Similarly VideoMAE \cite{videomae, videomaev2} reconstructs video through mask and achieved state-of-the-art results on multiple downstream tasks. There are also some other self-supervised pre-training tasks such as self-distillation \cite{dino, dinov2}, contrastive learning \cite{moco, cl, clr}, etc. In this work, we draws on \cite{AVT, unsupervised} to use the natural timing of the video to train the network and improve the reasoning ability of the network.

\section{Method}

The overall architecture of the ARR model is shown in \cref{fig:Architecture}. It is evident that this is a two-stage approach consisting of action recognition and sequence reasoning. We begin by feeding a sequence of video clips sampled from observed segments into the spatial-temporal encoder. Through action recognition supervision, this process enables the extraction of a feature sequence representing present-action information \cite{aim, timesformer}. Subsequently, this feature sequence is input to the subsequent causal decoder for NAP, facilitating the network's capacity to model the statistical relationships among actions. We will proceed to provide a detailed exposition of each component of the model, along with an introduction to training and implementation details.

\subsection{Sequence Modeling}

The first step we undertake is to sample video clips from the observed segments and model them as a sequence. Here, we sample $T$ clips at intervals of the anticipation time $ \tau_{a}$. Each clip comprises $n$ frames. Consequently, we obtain a video input $V$ in sequence form. Furthermore, we acquire the action category for each clip based on the original annotations of the dataset. For clips that lack annotations in the original dataset, we label them as unknown action (which will be discussed in the subsequent experimental section). This allows us to acquire action annotation sequences $A$ that correspond one-to-one with the input video clip sequences. We then employ these action annotation sequences as ground truth to provide supervision for subsequent action recognition and NAP tasks.

Modeling the input video as a sequence with strict temporal order is necessary. This necessity arises from the fact that the subsequent causal decoder needs to ensure that when predicting the next action, it exclusively focuses on the preceding action information and can not get information from the future to avoid information leakage. This approach enables robust modeling and representation of the statistical relationships and correlations among various actions. Such a rigorous configuration serves as the foundation for endowing ARR with formidable reasoning capabilities \cite{gpt2}.

\subsection{Action Recognition from Video Sequence}

The input of the network is a video sequence containing $T$ clips, $V = \left \{ C_{1}, C_{2},\cdots,C_{_{T} }   \right \} $, each clip contains $n$ frames, $C = \left \{ F_{1}, F_{2},\cdots,F_{_{n} }   \right \} $. Input $V$ into recognition Net, $\varphi$, it will process each clip separately, extract a feature $z$ for each clip for subsequent action classification and sent to the sequence reasoning network, the feature sequence can be expressed as $Z = \left \{ z_{1}, z_{2},\cdots,z_{_{T} }   \right \} $ where $z_{t} = \varphi(C_{t}) $. Similarly, each video clip $C_{t}$ has a corresponding action label $a_{t}$, the  action annotation sequence can be expressed as $A = \left \{ a_{1}, a_{2},\cdots,a_{_{T} }   \right \}$, which is in one-to-one correspondence with $V$.

\begin{figure}[thpb]
      \centering
      \includegraphics[width=0.6\linewidth]{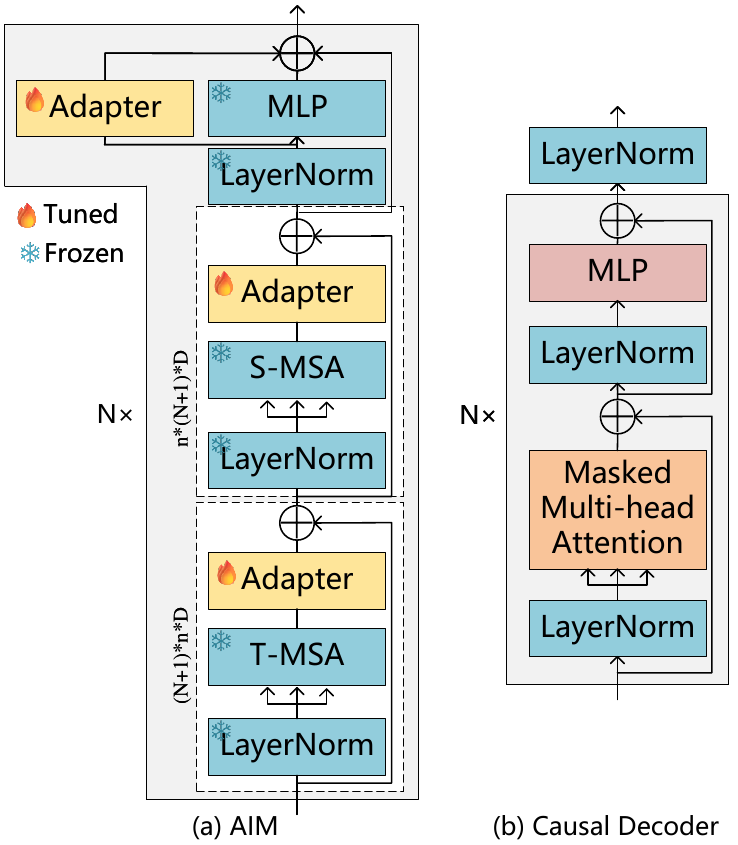}
      \vspace{-3mm}
      \caption{Architectures of AIM and Causal Decoder. (a) AIM uses vanilla ViT \cite{vit} model to initialize the weight of the blue part. Only the Adapter part of the model is trainable during training. (b) This is a standard decoder architecture with masked self-attention, which guarantees that the output is only related to the input before it, all parameters need to be updated during training.}
      \label{fig:aim_decoder}
      \vspace{-4mm}
   \end{figure}
   
The encoder part of ARR can use almost all video understanding models \cite{videomae, videoswin, timesformer, vivit}, and in this work we use AIM \cite{aim} as the backbone of our action recognition network to process clip-level input. The structure of AIM is shown in \cref{fig:aim_decoder}(a). AIM plugs a temporal processing module on the basis of ViT \cite{vit}, and achieves equal attention to space and time information through a clever transposition. It has a powerful spatial-temporal modeling ability and performs well in the action recognition task. AIM can easily process input video clips of arbitrary length, and when the input is a single frame, its performance is consistent with ViT, showcasing strong flexibility.

Specifically, a video $V$ containing $T$ clips is input to the encoder. The network independently processes each clip $C$, where $C\in \mathbb{R}^{n \times H \times W \times C} $, ensuring that their information remains non-interacting. Similar to most video-based models, we segment RGB frames of size $224 \times 224$ into non-overlapping patches of $16 \times 16$, these patches are then mapped into tokens of length $D$, generating $N$ tokens per frame. Additionally, a class token is appended for subsequent classification tasks. Consequently, the input clip $C$ becomes $n \times (N+1) \times D$ tokens. During temporal modeling, this layout is transposed to $(N+1) \times n \times D$ for temporal attention. Similarly, for spatial attention, it remains $n \times (N+1) \times D$. AIM processes these tokens to learn their spatial-temporal information. The final output is the average of $n$ class tokens, representing the clip, which accomplishes subsequent classification tasks and proceeds to the subsequent layers of the network. This yields a feature sequence $Z$ from $T$ input clips, where $Z\in \mathbb{R}^{T \times D} $.

Then, we classify the obtained action feature sequence $Z$ through a linear layer, supervised by the action label sequence $A$ obtained above, and we use the cross-entropy loss to calculate the first part of the action recognition loss:

\begin{equation}\label{eq1}
\mathcal{L}_{rec}=-\sum_{t=1}^{T} \log_{}{\hat{r}_{t} } [a_{t} ] ,
\end{equation}

\noindent where $\hat{r}_{t}$ is the classification result obtained by $z_{t}$ through the linear layer, and $a_{t}$ is the ground truth label.

The action recognition loss can ensure that the extracted features pay more attention to the current action information rather than other irrelevant information \cite{predicting} and is beneficial to the subsequent NAP task \cite{tempagg}.

\subsection{Sequence Reasoning from Action Feature Sequence}

The input to the sequence reasoning network is the previously generated action feature sequence $Z$, where each $z_{t}$ represents an action clip. When $Z$ is fed into the sequence reasoning network, $\psi$, it produces a new sequence $P$,
$P = \left \{ p_{1}, p_{2},\cdots,p_{_{T} }   \right \} $, where $p_{t} = \psi(z_{1}, z_{2}, \dots z_{t})$, We use $p_{t}$ to predict the next action, which only contains information from the past.

Similarly, the sequence reasoning module here can adapt to most sequence models, such as LSTM \cite{lstm}, GRU \cite{gru}, etc. Here we use the standard transformer causal decoder \cite{transformer, gpt2} based on the masked self-attention mechanism \cite{transformer}. As shown in \cref{fig:aim_decoder}(b), the structure of this pure decoder has been proved by many large language models to have a strong sequence reasoning ability through the NTP task, so here we use it as our sequence reasoning module.

\begin{algorithm}[t]
    \caption{Unsupervised Pre-training for ARR Decoder}
    \label{algo:1}
    \KwData{Input unlabeled long video $V$}
    \KwResult{Decoder $\psi$}
    Sample $n$ frames at regular intervals from $V$:
    ${f_{1}, \dots f_{n} } = Sample(V)$\
    
    Use Encoder $\varphi$ to extract features: 
    $\left \{ z_{1}, \cdots z_{_{n} }   \right \} =\varphi (\left \{ f_{1}, \cdots f_{_{n} }   \right \} )$\
    
    Use the decoder $\psi$ to predict the next feature:
    $\left \{ \hat{z}_{2}, \cdots \hat{z}_{n+1}    \right \} =\psi (\left \{ z_{1}, \cdots z_{_{n} }   \right \} )$\
    
    Calculate the loss using $MSE \ loss$:
    $loss = \sum_{t=2}^{n} MSE(\hat{z}_{t}, {z}_{t})$ \
    
    Compute gradients and update parameters:
    $loss.backward(),$
    $\, update(\psi.params())$ \
    
    \Return{$\psi$}
\end{algorithm}

The masked self-attention mechanism is the key to the network's ability to predict the future \cite{ntp}. It uses the observed action information to predict the next action, and effectively models the correlation and statistical relationship between actions. Specifically, our predicted next action feature $p_{t}$ goes through a linear layer for classification, and uses the next action label $a_{t+1}$ to supervise it, thus obtaining the prediction loss:

\begin{equation}\label{eq2}
\mathcal{L}_{pre}=-\sum_{t=1}^{T} \log_{}{\hat{y}_{t} } [a_{t+1} ] ,
\end{equation}

\noindent where $\hat{y}_{t}$ is the classification result obtained by $p_{t}$ through the linear layer, and $a_{t+1}$ is the ground truth label, here we also use the cross entropy loss function.

The prediction loss can ensure that our sequence reasoning network focuses on the correct judgment of the next action \cite{predicting}. This training method enables the enhanced network to learn the correlation between actions, thereby improving the prediction ability \cite{gpt2}.

\subsection{Unsupervised Pre-training based on Video}

\begin{table}[t]
\centering
\caption{Experiments to verify action associations. Three labeling methods were tested for the clips that were not marked at the corresponding time. “Unknown” refers to adding an unknown action that does not exist in the data set to label, “Random” refers to randomly selecting an existent action to label. “Previous” refers to using the previous action to label. length refers to the length of the input sequence. cm Recall refers to class mean Top-5 Recall. The best results are highlighted in \textbf{bold}.}
\label{table1}
\resizebox{\linewidth}{!}{\begin{tabular}{c|ccccc|ccccc|ccccc}
\hline
Method    & \multicolumn{5}{c|}{Unknown}  & \multicolumn{5}{c|}{Random}   & \multicolumn{5}{c}{Previous}  \\ \hline
Length    & 4     & 6     & 8     & 10  & 12   & 4     & 6     & 8     & 10  & 12    & 4     & 6     & 8     & 10   & 12 \\ \hline
cm Recall  & 16.04  & 17.51 & \textbf{19.09} & 17.93 & 16.95 
 & 15.27  & 15.64 & 15.51 & 16.48 & 16.19
 & 16.69  & 16.79 & 17.38 & 17.02 & 17.23
\\ \hline
\end{tabular}}
\end{table}

As we all know, the ability of large language models based on causal decoders is not good enough at the beginning \cite{gpt1, gpt2}, and its performance has been improved qualitatively with the large increase in the amount of training data \cite{gpt2, gpt3}. However, in the task of action anticipation, labeled videos only accounting for a small fraction, most videos are still unlabeled, so we need to find an unsupervised pre-training method to improve network performance. However, the vanilla NTP task cannot be directly applied to video pre-training as video frames are not limited in quantity like words, and we cannot build a vector for each frame to represent it \cite{transformer, anticipating}.

Drawing on the work \cite{anticipating, unsupervised}, we propose a method that leverages the natural timing of videos for unsupervised pre-training. As shown in Algorithm 1, we do not directly predict the next video frame, but predicting the features of the next video frame through a specific encoder. We use the distance loss like $MSE \ loss$ to force the causal decoder to learn a specific representation of the next frame. In this way, the reasoning ability of the decoder can be enhanced. With this unsupervised pre-training method, almost all long videos can be used for training, which greatly expands the scale of trainable data.

\subsection{Total Loss}

We train ARR end-to-end with the above recognition loss and prediction loss together for supervision:

\begin{equation}\label{eq3}
\mathcal{L}_{total} = \mathcal{L}_{rec} +\mathcal{L}_{pre} .
\end{equation}

\noindent The recognition loss and prediction loss here are supervised by the same action label sequence and cross entropy loss, but there is a shift in time, which is the key to the NAP task, noted that the action labeling sequence requires us to pre-process the annotation file of the dataset to obtain.

\begin{table}[t]
\centering
\caption{Comparison of state-of-the-art methods on the validation set of EK100. All mothods only use RGB modality as input to be fair, and the evaluation index is class mean Top-5 Recall. The best results are highlighted in \textbf{bold}.}
\label{table2}
\begin{tabular}{llll}
\hline
Method & Verb             & Noun             & Action \\ \hline
RULSTM \cite{RULSTM} & 27.5             & 29.0             & 13.3   \\
TemAgg \cite{tempagg} & 23.2            & 31.4            & 14.3  \\
AVT \cite{AVT}   & 30.2             & 31.7             & 14.9   \\
AFFT  \cite{AFFT} & \textbackslash{} & \textbackslash{} & 16.1   \\
ARR (Ours)    & \textbf{31.3}             & \textbf{32.0}             & \textbf{16.3}   \\ \hline
\end{tabular}

\end{table}

\section{Experiments}

\subsection{Experimental Setup}
This section discusses the experimental related settings, including the datasets used for evaluation and evaluation measures.

\textbf{Datasets.} We conducted experiments on three large-scale datasets: Epic-kitchen-100 (EK100) \cite{EK100}, EGTEA Gaze+ \cite{gteadataset}, and 50salads \cite{50salads}. EK100 is an unscripted egocentric action dataset collected from 45 kitchens in four cities worldwide. It includes $100$ hours of video content, with $20M$ frames extracted and annotated with $9K$ action segments. The dataset comprises $97$ verb classes and $300$ noun classes, which is the largest dataset in the field of action anticipation tasks. Epic-kitchen also has an early version Epic-kitchen-55 (EK55) \cite{EK55}, which contains $55$ hours of video, EK100 contains EK55. The anticipation time in EK100 is set at $1s$. EGTEA Gaze+ is an egocentric video dataset containing $10K$ action annotations, covering $19$ verbs, $51$ nouns, and $106$ unique actions. The anticipation time is set at $0.5s$. 50salads is a smaller-scale third-person perspective video dataset, consisting of $0.9K$ action segments, encompassing 17 different actions, we use it to demonstrate the generality of our method, the anticipation time is set at $1s$.

\textbf{Metrics.} Referring to previous work \cite{AVT, RULSTM, AFFT}, we employ the Top-k metric to evaluate our method, as it fundamentally remains a classification task. That is, if the top-k predictions include the correct ground truth action label, we consider the prediction to be accurate. For categories with a substantial number of classes, such as EK100, we use the class mean Top-5 recall (cm Recall) metric, obtained by averaging the Top-5 recall values across all classes. As observed in prior studies \cite{anticipatinghuman, leveraging}, considering the uncertainty of future predictions (i.e., many plausible actions can be taken after observation), this evaluation approach is suitable. Averaging across all categories requires the network to equally prioritize each class, rather than disregarding classes with fewer instances. The task of action anticipation is relatively challenging, and if using Top-1 accuracy on datasets would yield markedly low results, which wouldn't effectively gauge the network's performance. All results in the table are displayed in percentage form, and $\%$ is omitted.

\begin{table}[t]
\centering
\caption{Comparison of state-of-the-art methods on the EGTEA Gaze+. Methods marked with * are averaged across the three official splits, while others are based on split 1 only. Top-1 refers to top1 accuracy, cm Top-1 refers to class mean Top-1 accuracy. cm Top-1 needs to take into account the balance between classes more. The best results are highlighted in \textbf{bold}.}
\label{table3}
\begin{tabular}{cccc|ccc}
\hline
\multirow{2}{*}{Method} & \multicolumn{3}{c|}{Top-1}  & \multicolumn{3}{c}{cm Top-1} \\ \cline{2-7} 
                        & Verb & Noun & Action        & Verb  & Noun & Action        \\ \hline
I3D-Res50 \cite{quo}               & 48.0 & 42.1 & 34.8          & 31.3  & 30.0 & 23.2          \\
FHOI \cite{FHOI}                   & 49.0 & 45.5 & 36.6          & 32.5  & 32.7 & 25.3          \\
AVT  \cite{AVT}                   & \textbf{54.9} & \textbf{52.2} & 43.0          & \textbf{49.9}  & \textbf{48.3} & \textbf{35.2} \\
AFFT* \cite{AFFT}                & 53.4 & 50.4 & 42.5          & 42.4  & 44.5 & \textbf{35.2} \\
ARR (Ours)                & 53.5 & 52.1 & \textbf{43.1} & 47.5  & 46.3 & 35.0          \\ \hline
\end{tabular}

\end{table}

\begin{table}[t]
\centering
\caption{Comparison of state-of-the-art methods on 50salads. Because there are few types of actions, the Top-1 accuracy is directly used as the evaluation indicator here. ARR outperforms prior work in 50salads dataset. The best results are highlighted in \textbf{bold}.}
\label{table4}
\begin{tabular}{c|c}
\hline
Method      & Top-1         \\ \hline
RNN\cite{cnnrnn}         & 30.1          \\
CNN\cite{cnnrnn}         & 29.8          \\
ActionBanks\cite{temporal} & 40.7          \\
AVT\cite{AVT}         & 48.0          \\
ARR (Ours)       & \textbf{49.2} \\ \hline
\end{tabular}

\end{table}
\subsection{Implementation Details}
Similar to most experimental settings in \cite{AVT, keeping, omnivore}, we first change the height of all input RGB images to $256$, and then crop them to $224 \times 224$ during training. During testing, we utilize a $3$-crop approach, wherein we calculate three spatial crops in input frames of size $224$. Subsequently, we average the predictions over these three corresponding inputs. In most experiments, $8$ clips are sampled at intervals of anticipation time, and each clip contains $4$ frames. 
The encoder part uses the AIM pre-training model. 
The decoder, as shown in \cref{fig:aim_decoder}(b), is configured to match the size of the GPT-2 Small model \cite{gpt2}. Specifically, it comprises 12 Transformer blocks, with 12 heads, and a model dimension of 768. It has a total of approximately 120M parameters and a computational load of 2.3 GFlops.
It is initialized using the above-mentioned unsupervised pre-training. We also tried the GPT2 pretrained model, and their depth $N$ is set to $12$. We train ARR end-to-end with Adam using $4*10^{-5}$ weight decay and $10^{-4}$ learning rate for $50$ epochs, with a $20$ epochs warmup \cite{warmup} and $30$ epochs of cosine annealed decay \cite{cosinelr}. We set the batch size to 8 and train the model on two 3090 GPU for approximately 50 hours.

\subsection{Statistical correlation between actions}

Firstly, we conducted experiments to verify the statistical correlations between actions. We directly employed the action annotation sequence $A$ mentioned earlier for validation. Training was performed on sequence $A$ obtained from the training set, and performance testing was carried out on the validation set. The causal decoder mentioned earlier was utilized as the network architecture. After embedding the action labels, they were fed into the network. The evaluation was based on the same class mean Top-5 recall metric as previously mentioned. 

It can be seen from \cref{table1} that only using action sequences for reasoning can achieve good performance, indicating that there is indeed a strong statistical relationship among actions, which verifies our conjecture. The best performance is achieved when the input sequence length is $8$ and unknown action class is introduced to populate unannotated segments. Too short and too long input lengths are inappropriate. When the input length is too short, the input information is insufficient to effectively model the potential correlation. When the input length is too long, the correlation between the previous information and the subsequent information is too weak to introduce useless information and increase the amount of calculation, resulting in greater redundancy \cite{lamda}. This setup is maintained throughout subsequent experiments. It is worth noting that these experiments are only to confirm the statistical correlation between actions and cannot be used to measure ARR network performance for it does not use video input.

\subsection{Comparison to the state-of-the-art}
 
\noindent\textbf{EK100.} We first compare ARR with previous work \cite{AFFT, AVT, FHOI, quo} on the validation set of EK100, since our input only takes RGB single modality input. For a fair comparison, only those single modality methods are compared. As shown in \cref{table2}, our method achieves the best performance, surpassing the previous best work AVT, also based on the encoder-decoder architecture. The architectures of these models are similar, but our method better models the correlation between actions instead of just recursively predicting some frame or clip-level representations and makes better use of this prior information.

It can be seen that the performance of our method is approaching the results in \cref{table1}. Although the action recognition part will cause some performance differences, there are some additional scene information in the video that we can capture to make up for part of the gap.

\vspace{1mm}
\noindent\textbf{EGTEA Gaze+.} Next we evaluate our method on EGTEA Gaze+, as shown in \cref{table3}. Following previous work \cite{RULSTM}, we set the expected time $\tau_{a}$ to be $0.5s$. It can be seen that our method can also achieve good performance in different anticipation time settings, which proves that our method is robust and can be used in a variety of anticipation task settings. Because the number of action categories is small, we report the Top-1 accuracy and class mean Top-1 accuracy respectively, the latter is averaged between each category regardless of the number of samples in each category, requiring the network to pay equal attention to each category instead of ignoring those classes with a small number of samples.

\vspace{1mm}
\noindent\textbf{50salads.} Finally, we show that ARR is not limited to first-person videos, but also works in third-person videos. In contrast to first-person videos that primarily focus on the hand area, in third-person videos, the person's body is also largely visible. We report Top-1 accuracy averaged over the pre-defined $5$ splits following prior work \cite{temporal} in 
\cref{table4}, because 50salads has a small number of action categories, this indicator can be used for evaluation. We can see that our method outperforms previous methods and can achieve an accuracy close to $ 50 \%$, demonstrating the generality of our method.

\subsection{Ablations and Analysis}
   
\begin{table}[t]
\centering
\caption{Ablations on unsupervised pretraining. We conduct a comparison test on the EK100 validation set, and compare the decoder obtained by our unsupervised pre-training with other methods. “our pretrain” refers to pre-training on the EK100 dataset, which is also the default practice in this paper, “pretrain (on EK100 and K400)” refers to pre-training on the EK100 dataset and K400 \cite{k400} dataset. The encoder part uses AIM, and the evaluation indicators are cm Recall. The best results are highlighted in \textbf{bold}.}
\label{table5}
\begin{tabular}{c|ccc}
\hline
Decoder         & Verb          & Noun          & Action        \\ \hline
GPT2 \cite{gpt2}            & 30.8          & 32.1 & 16.0          \\
w/o pretrain & 27.9          & 29.6          & 14.1          \\
our pretrain & 31.3 & 32.0          & 16.3 \\ 
pretrain (on EK100 and K400) &\textbf{31.6}  &\textbf{32.3}  &\textbf{16.5} \\ \hline
\end{tabular}

\end{table}

\noindent\textbf{The effect of unsupervised pre-training.} We first investigated whether our proposed unsupervised pretraining method using the natural temporal order of videos is effective for the causal decoder. We employed the ARR architecture, where the action recognition network utilized ARR. The sequence inference networks were implemented with three different decoders: GPT2 decoder, randomly initialized decoder, and the decoder obtained through our method. We compared their performance on the EK100 validation dataset.

\begin{figure}[t]
      \centering
      \includegraphics[width=0.9\linewidth]{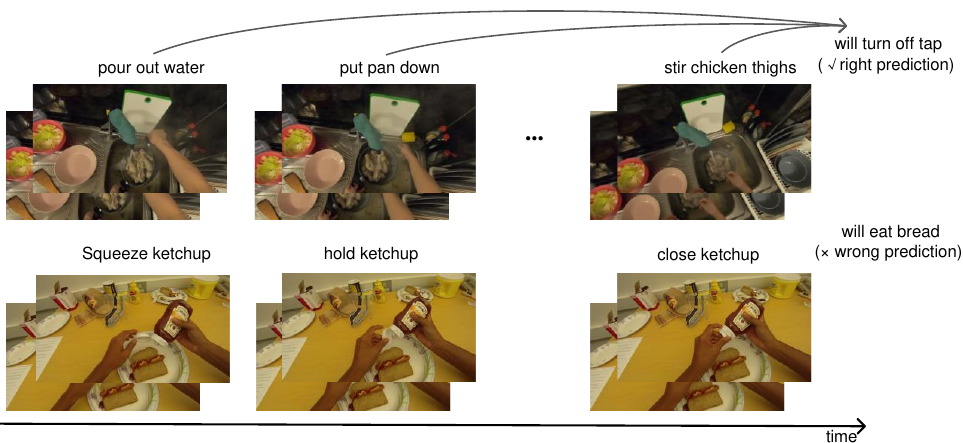}
      \vspace{-3mm}
      \caption{Schematic diagram of inference process. ARR first judges the current action, and then judges the actions that may occur next according to the current action sequence. We demonstrate one correct inference and one incorrect inference.}
      \vspace{-4mm}
      \label{fig:inference}
   \end{figure}
   
As shown in \cref{table5}, we can see that our method achieves the best results. GPT2 \cite{gpt2} is a large language model that uses text pre-training and is also used in reasoning networks for other work. It has been proven to have a strong reasoning ability. The method of random initialization has poor performance, which is very reasonable, because it is difficult to obtain good performance directly through fine-tuning for a large-scale network without large-scale pre-training \cite{mae}. Our method uses video data for pre-training, although there are not as many training data as GPT2, but the performance is comparable or even slightly exceeded. We guess that the performance of GPT2 may not be fully utilized because of the domain gap between text pre-training and video pre-training. In other words, the result can prove the effectiveness of our unsupervised pre-training method.

In addition, we further expanded the size of the pre-training data. It can be seen that when the size of the pre-training data is expanded from the EK100 dataset to the EK100 dataset and the K400 dataset \cite{k400}, the performance of the network is further improved, which shows that our pre-training method can make the performance of the decoder further improves with the increase of pre-training data, demonstrating the scaling potential of our unsupervised pre-training method.
\begin{table}[t]
\centering
\caption{Ablations on different encoders. We conduct a comparison test on the EK100 validation set. The decoder part uses GPT2, and the evaluation indicators are cm Recall.}
\label{table7}
\begin{tabular}{c|ccc}
\hline
encoder         & Verb          & Noun          & Action        \\ \hline
Timesformer \cite{timesformer}            & 30.8          & 31.9 & 16.0          \\
Vivit \cite{vivit} & 30.5         & 31.4          & 15.8          \\
VideoMAE \cite{videomae} & 31.0 & 32.0          & 16.1 \\ 
AIM \cite{aim} &31.3   &32.0    &16.3
\\ \hline
\end{tabular}

\end{table}
\begin{table}[t]
\centering
\caption{Effect of clip length on performance. We explored the impact of the length of a single clip on the performance of the model. We compared the performance on the EK100 verification set. }
\label{table6}
\begin{tabular}{c|ccccc}
\hline
Num\_frames & 1    & 2    & 4    & 6    & 8    \\ \hline
cm Recall   & 14.8 & 15.2 & 16.3 & 16.5 & 16.8 \\ \hline
\end{tabular}

\end{table}

\begin{figure}[t]
      \centering
      \includegraphics[width=0.9\linewidth]{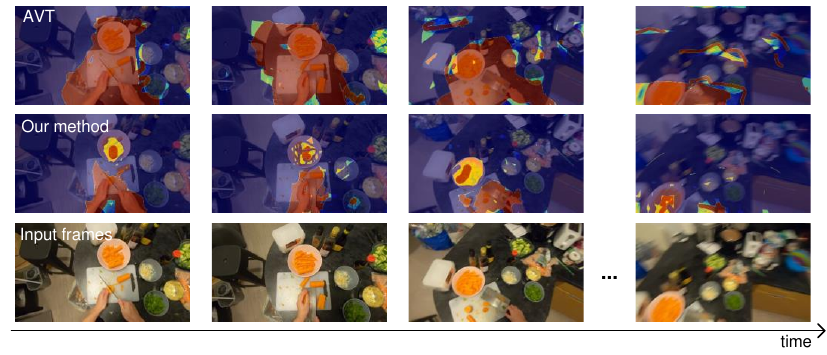}
      \vspace{-3mm}
      \caption{Attention visualization. ARR is able to automatically focus on regions that are more relevant to actions and the scope is more precise.}
      \vspace{-4mm}
      \label{fig:attention}
   \end{figure}
\vspace{1mm}

\noindent \textbf{The effect of different encoders.} As shown in Table~\ref{table7}, we use different action recognition networks as encoders to verify the effectiveness of our ARR architecture. It can be seen that the performance of using these mainstream action recognition networks as encoders is very good. Among them, using AIM as the encoder has the best performance, and only needs to adjust the parameters of the adapter part, so in this work we use AIM as the encoder by default.

\vspace{1mm}
\noindent\textbf{Clip Length.} In \cref{table6}, we explore the impact of the number of frames contained in a single clip in the input video sequence on performance. It can be seen that as the number of frames increases, the performance of the network gradually improves. This is easy to understand. In the action classification task, as the number of input frames increases, the performance of the network will also improve. However, it can be seen that when the number of frames exceeds 4, the performance improvement is not so obvious. In consideration of the tradeoffs between performance and calculation, we use 4 frames as the length of the clip in most experiments.

\vspace{1mm}
\noindent\textbf{Predicting Next Action.} As shown in \cref{fig:inference}, we show the inference process of ARR. ARR will recognize the current action based on the current action state, first “pour out water”, then “put pan down”, “stir chicken thighs”, ARR will judge that the next action is “turn off tap” based on the existing action state combined with the correlation between the actions it has learned. Thus, this way of reasoning is consistent with what we expected. 

However, this inference mode will also bring some errors. When the actions are too dense, ARR will make wrong judgments, as shown in the bottom part of \cref{fig:inference}. The true label is to continue to “close ketchup”, while the prediction given by ARR is “eat bread”. ARR tends to be judged based on experience, but it is easy to ignore the speed of the action.

\vspace{1mm}
\noindent\textbf{Visualization.} We analyze models' visual attention of the last layer of the encoder on the EK100 dataset to qualitatively evaluate the effectiveness of learnable spatio-temporal representation. As shown in \cref{fig:attention}, it is clear that through the supervision of the action recognition task, the network can automatically focus on the areas that are more relevant to the current action (usually the hand area). Although there are many objects in the frame, the model still learns to pay more attention to the objects related to the current action instead of others irrelevant. Compared to other methods, our focus areas are also more precise and specific.

\addtolength{\textheight}{-3cm}   
                                  

\section{CONCLUSIONS}   
We proposed ARR, an end-to-end video modeling architecture based on attention mechanisms. By decomposing the action anticipation task into action recognition and sequence reasoning tasks, we constructed predicting the next action through a temporal shifting supervision, which effectively modeled and learned the underlying relationships among actions and achieved competitive results on multiple datasets. Our proposed method of utilizing the natural temporal order of videos for unsupervised pre-training also resolved the problem of NAP requiring large number of training data. We hope to continue to explore unsupervised training methods related to action anticipation in the future.
\section{Acknowledgments}

This work was supported in part by the National Natural Science Foundation of China under Grant 62171309 and 62306061, Guangdong Basic and Applied Basic Research Foundation (Grant No. 2023A1515140037), and Open Fund of National Engineering Laboratory for Big Data System Computing Technology (Grant No. SZU-BDSC-OF2024-02).








\bibliographystyle{ACM-Reference-Format}
\bibliography{ref}










\end{document}